# Feature-Optimized Vision for Adaptive 3D Scene Reconstruction

**Eric Liang**
Oracle
zixuan.liang@oracle.com

**Abstract.** Three-dimensional scene reconstruction depends on local image evidence that is both visually discriminative and geometrically useful. Fixed feature thresholds and uniform feature budgets are easy to deploy, but they can waste computation on repeated texture, low-parallax regions, or unstable points. This paper proposes an adaptive feature-optimized vision front end for 3D reconstruction. The method scores candidate features by texture, repeatability, distinctiveness, expected triangulation angle, and spatial coverage, then allocates a per-view feature budget to maximize useful tracks under a fixed reconstruction pipeline. A small synthetic multi-view prototype evaluates four selection policies across corridor, facade, object-table, and cluttered scenes. Compared with random, texture-only, and uniform-grid baselines, the adaptive policy obtains the best quality-aware completeness and the lowest aggregate reconstruction RMSE while preserving broad image coverage. The result is not a replacement for modern learned matching or neural reconstruction systems; it is a modular front-end policy that can make classical and learned 3D pipelines more deliberate about which visual evidence they spend compute on.

## 1. Introduction

3D scene reconstruction pipelines turn overlapping images into camera poses, sparse tracks, point clouds, meshes, or neural scene representations. Even when the back end is sophisticated, the front end still depends on a basic question: which image observations are worth trusting and matching? A reconstruction can fail not because there are too few pixels, but because the selected features are repetitive, unstable, weakly triangulated, or concentrated in one part of the image.

Classical systems such as SIFT-based structure from motion, ORB-based real-time mapping, and COLMAP-style sparse reconstruction rely on hand-engineered or learned feature detectors, descriptors, robust matching, and geometric verification. Newer learned matchers and 3D-aware models improve correspondence quality, yet practical systems still make budget decisions: how many features to extract, where to place them, when to drop uncertain matches, and how to balance coverage with distinctiveness.

This paper studies feature selection as an adaptive control problem for 3D reconstruction. The proposed front end estimates a feature quality field over each view and uses that field to allocate a constrained feature budget. The policy favors features that are locally discriminative, repeatable across likely viewpoints, spatially diverse, and expected to yield useful triangulation angles. The same idea can sit in front of classical sparse SfM, multi-view stereo initialization, Gaussian-splatting bootstrapping, or learned matching systems.

The contribution is a compact algorithmic formulation, a reusable reconstruction architecture, and a synthetic prototype that exposes tradeoffs among feature quality, coverage, geometric accuracy, and compute. The prototype is intentionally modest: it is not a benchmark against production systems or a claim of state-of-the-art reconstruction. Its purpose is to isolate the role of adaptive feature optimization under controlled scene conditions.

## 2. Background and Related Work

Local feature methods remain a foundation of image-based reconstruction. Lowe's SIFT detector and descriptor established scale- and rotation-invariant matching as a robust basis for object recognition and reconstruction [1]. RANSAC supplied the robust estimation pattern that lets geometric models tolerate outliers [2]. FAST and ORB made corner detection and binary descriptors practical for fast matching and real-time systems [3], [4]. These methods are computationally efficient and interpretable, but they often depend on thresholds and spatial heuristics that are not explicitly optimized for downstream 3D geometry.

Multi-view geometry and structure from motion provide the reconstruction substrate. Hartley and Zisserman's treatment of multiple-view geometry remains the standard reference for epipolar constraints, triangulation, and bundle adjustment [5]. COLMAP showed how carefully engineered feature extraction, matching, incremental mapping, and bundle adjustment can produce robust general-purpose SfM [6]. Multi-view stereo then densifies sparse geometry by exploiting photometric consistency across calibrated views [7].

Learned local features and matchers have changed the quality frontier. SuperPoint learns interest points and descriptors through self-supervision [8]. R2D2 emphasizes repeatability and reliability [9]. SuperGlue uses graph neural networks and attention for context-aware matching [10], while LoFTR removes the explicit detector stage and matches dense local features with transformers [11]. LightGlue further improves the speed-adaptivity tradeoff for local feature matching [12]. These systems show that feature quality is not a scalar texture measure; it combines local appearance, context, confidence, and matchability.

Neural and 3D-aware scene representations add another layer. NeRF represents scenes as radiance fields optimized from posed images [13], and 3D Gaussian Splatting offers real-time rendering from optimized Gaussian primitives [14]. DUSt3R and MASt3R point toward direct 3D reasoning from image pairs or collections [15], [16]. Even in these newer families, front-end evidence selection matters because initialization quality, image overlap, uncertainty, and computational budget shape the final reconstruction.

Adjacent work on data representation, metadata organization, robustness estimation, and data cleaning informs the broader engineering context for reconstruction systems. High-cardinality categorical representations are relevant when feature vocabularies, scene identifiers, or learned codebooks become large [17]. Metadata harmonization matters for organizing image collections, calibration records, and reconstruction outputs [18]. Certified-radius estimation highlights the value of explicit uncertainty calculations in vision systems [19]. Automated date-format detection illustrates a recurring engineering problem in reconstruction workflows as well: noisy input metadata must be converted into reliable structured semantics [20].

## 3. Problem Formulation

Consider a set of images $I_1...I_n$ with known or estimated camera intrinsics and approximate view overlap. A feature candidate f in image i has image location x, local appearance descriptor d, detector response t, reliability estimate r, distinctiveness estimate q, and predicted triangulation usefulness p with respect to neighboring views. A reconstruction front end must choose a subset $S_i$ under a feature budget B per image.

A fixed detector threshold treats this as a local image problem. An adaptive reconstruction front end treats it as a downstream utility problem. A useful feature is not merely strong in one image; it should be matchable, geometrically informative, not redundant with nearby features, and not obviously unstable. The feature utility used in this prototype is a weighted score $s(f)=0.28t+0.27r+0.22q+0.19p-0.20a$, where a estimates ambiguity or repetition. Spatial caps prevent all selected features from collapsing into a single textured region.

The optimization is deliberately lightweight. It is not a full bundle-adjustment objective and does not require differentiating through the reconstruction back end. Instead, it supplies a practical feature-budget policy: rank candidates by expected reconstruction utility, enforce coverage constraints, reconstruct, measure diagnostics such as inlier ratio and triangulation angle, and update the next selection policy.

## 4. Adaptive Feature-Optimized Reconstruction

Figure 1 shows the proposed pipeline. Images first produce a feature quality field, either from classical detector statistics, learned confidence maps, or hybrid cues. The adaptive budget layer selects features that balance local quality and global coverage. Matching and geometric verification then estimate tracks, camera relations, and triangulated structure. Diagnostics feed back into future thresholds and regional quotas.

Adaptive feature-optimized 3D reconstruction pipeline
Image sequence
calibrated or estimated views with changing baseline and overlap
Feature quality field
texture, repeatability, distinctiveness, parallax, spatial coverage
Adaptive budget
select features that maximize expected stable tracks under fixed compute
3D reconstruction
robust matching, pose graph, triangulation, dense back end
Diagnostics
inlier ratio, triangulation angle, coverage, residual error
Policy update
adjust thresholds and regional quotas for next batch
The same scoring layer can feed sparse SfM, MVS initialization, or Gaussian-splatting/radiance-field bootstrapping.

*Figure 1. Adaptive feature-optimized 3D reconstruction pipeline.*

### 4.1 Feature Quality Components

The scoring layer combines five signals. Texture measures whether the local patch is informative. Repeatability estimates whether the feature will remain detectable under viewpoint, illumination, and scale changes. Distinctiveness penalizes repeated patterns that create plausible but wrong matches. Parallax usefulness estimates whether the observation can support stable triangulation rather than a near-degenerate ray intersection. Coverage keeps the selected feature set distributed across the image so pose estimation does not depend on one local surface.

### 4.2 Adaptive Budget Allocation

The selection algorithm first computes candidate scores, then applies a soft per-tile cap. This differs from a uniform grid, which guarantees coverage but may keep low-quality features, and from a texture-only policy, which may concentrate on repeated or planar texture. The adaptive policy keeps the implementation simple enough to plug into existing pipelines while still reflecting downstream geometric utility.

### 4.3 Reconstruction Back End

The prototype uses robust track formation and triangulation rather than a full bundle-adjustment system. In a production implementation, the same selected features would flow into established SfM or SLAM machinery: descriptor matching or learned matching, essential/fundamental matrix estimation, pose graph construction, triangulation, local and global bundle adjustment, and optionally multi-view stereo or neural scene optimization.

## 5. Prototype and Evaluation

The prototype simulates four scene types: an indoor corridor with weak texture and small baseline risk, a repetitive facade, an object-table scene with strong local structure, and cluttered vegetation with unstable features. Six calibrated views observe synthetic 3D points. Each candidate point receives texture, ambiguity, repeatability, distinctiveness, and visibility attributes. Four feature policies are compared under the same per-view budget: random selection, texture-only selection, uniform-grid selection, and the adaptive optimized policy.

The evaluation reports reconstruction RMSE, quality-aware completeness, inlier ratio, spatial coverage, and a runtime proxy. Quality-aware completeness counts reconstructed points whose simulated error is below 0.10 scene units; this avoids rewarding noisy point clouds that contain many unreliable tracks. The runtime proxy includes feature scoring, matching effort, and robust-estimation difficulty; it is not wall-clock performance.

Figure 2 gives a qualitative view of the synthetic setup. The prototype renders two camera observations of a procedurally generated table-top scene, overlays features selected by the adaptive budget policy, and shows the sparse 3D structure recovered from matched feature tracks. This figure is generated data, not an external image dataset.

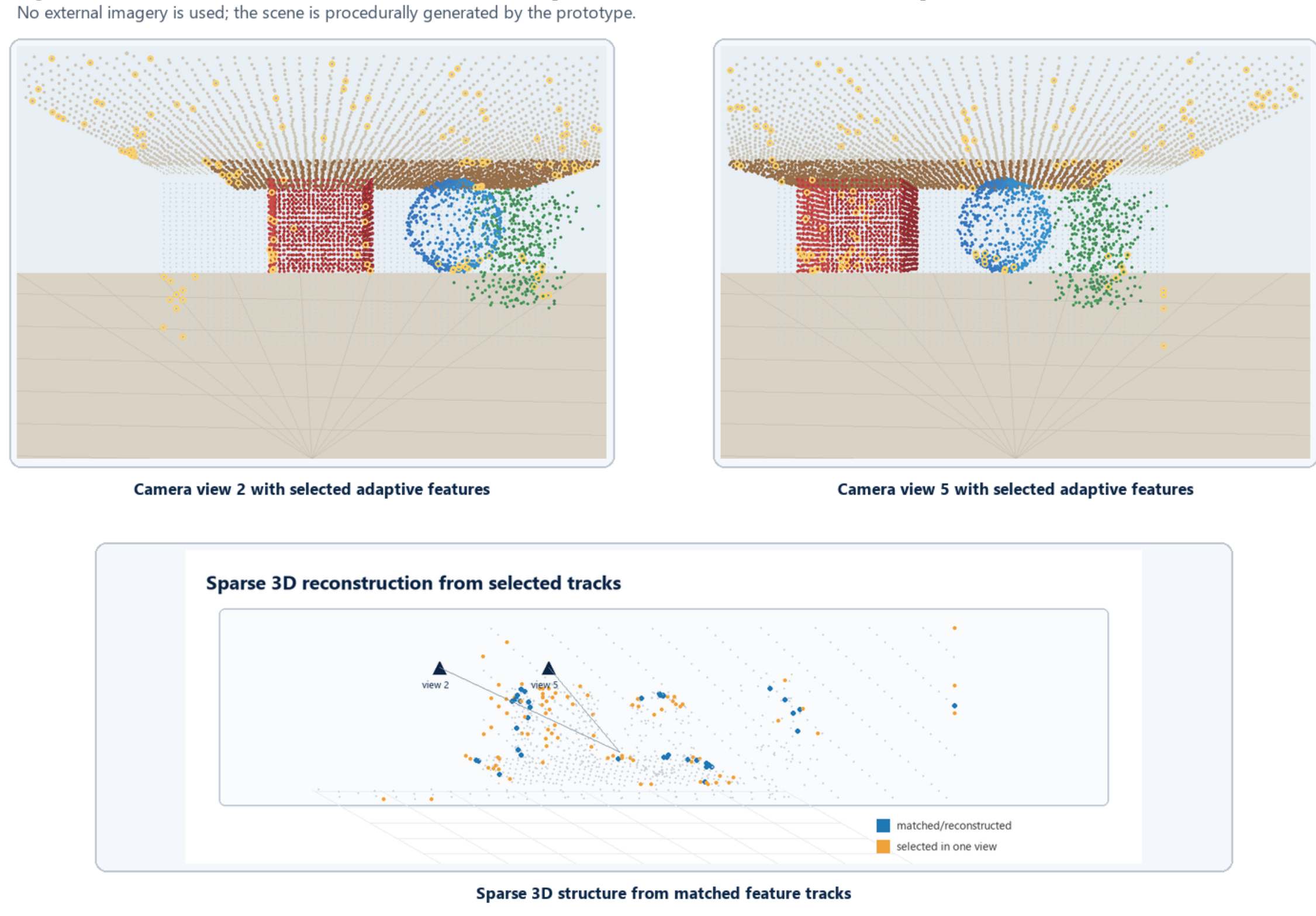

*Figure 2. Synthetic multi-view scene, adaptive feature overlays, and sparse 3D reconstruction output generated by the prototype.*

| Strategy | RMSE | Acc. compl. % | Inlier % | Coverage % | Runtime |
|---|---|---|---|---|---|
| Random budget | 0.213 | 10.8 | 65.1 | 96.2 | 38.4 |
| Texture-only | 0.087 | 20.3 | 80.5 | 96.2 | 39.2 |
| Uniform grid | 0.107 | 18.9 | 76.7 | 100.0 | 40.7 |
| Adaptive optimized | 0.085 | 21.6 | 80.7 | 97.5 | 42.6 |

*Table 1. Aggregate prototype results across four synthetic scene types.*

The adaptive policy achieves the best aggregate RMSE, best quality-aware completeness, and best inlier ratio. Uniform-grid selection provides the broadest spatial coverage, but it retains weaker points in low-quality regions. Texture-only selection is competitive on RMSE, especially in scenes where high texture is also distinctive, but it is more vulnerable to repeated facades and coverage concentration. Random selection sometimes reconstructs many points, but its higher error and lower inlier ratio make those points less useful.

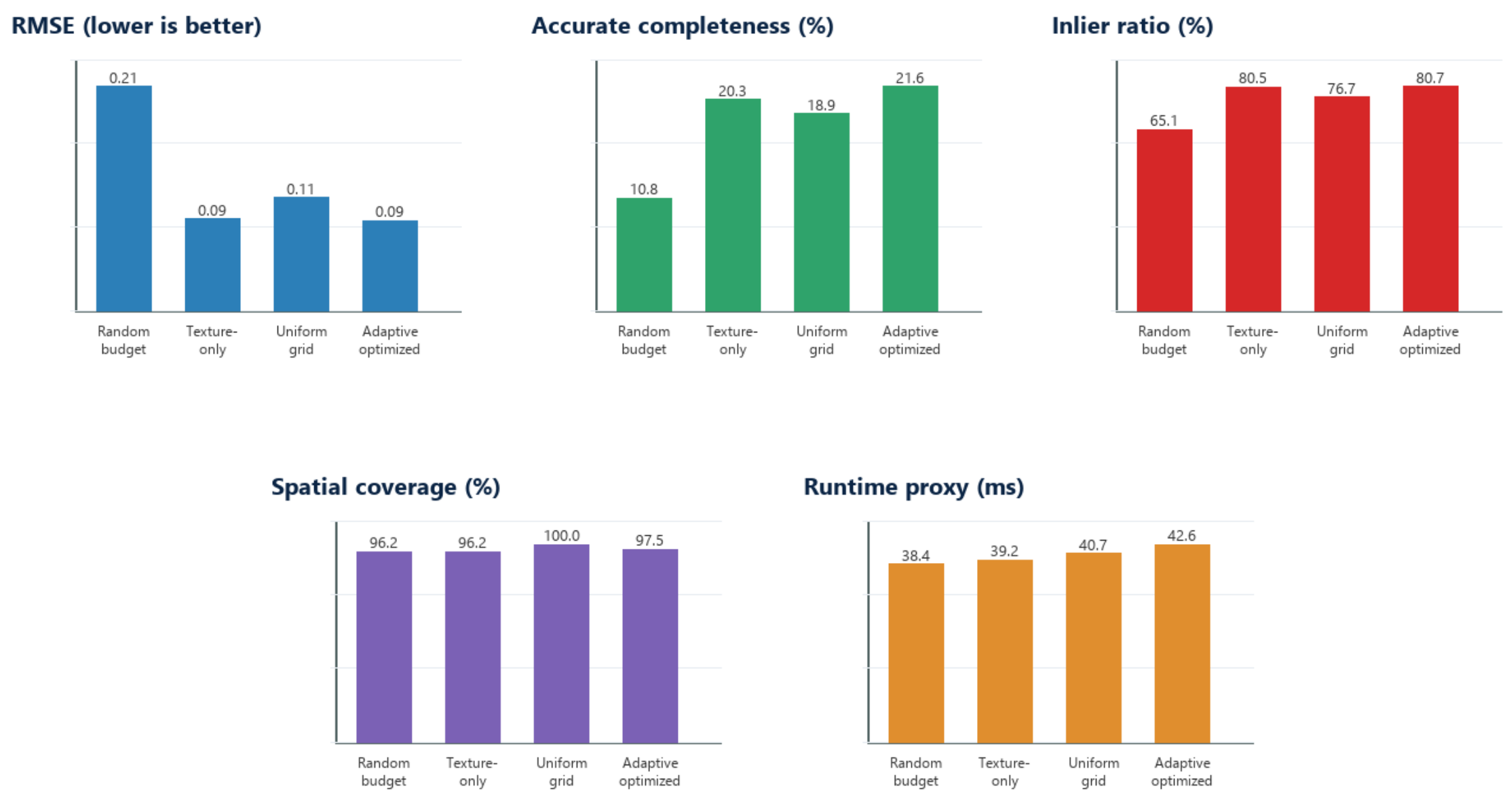

*Figure 3. Aggregate synthetic reconstruction metrics across four scene types.*

| Scene | Adaptive RMSE | Best baseline RMSE | Adaptive acc. compl. % | Best baseline acc. compl. % |
|---|---|---|---|---|
| indoor corridor | 0.136 | texture: 0.084 | 14.7 | uniform: 15.3 |
| facade repetition | 0.067 | texture: 0.098 | 14.6 | texture: 14.5 |
| object table | 0.051 | texture: 0.057 | 31.4 | texture: 29.4 |
| vegetation clutter | 0.086 | texture: 0.109 | 25.6 | texture: 22.4 |

*Table 2. Scene-level comparison of the adaptive policy against the best baseline.*

## 6. Discussion

The results support a conservative claim: adaptive feature optimization can improve the usefulness of a fixed feature budget by considering matchability and geometry together. The most important metric is not the number of detected features, but the number of reliable, well-triangulated, spatially distributed tracks that survive geometric verification. This aligns with modern learned matchers, which increasingly expose confidence and context rather than treating all local evidence equally.

The method is also useful as an engineering pattern. It separates the feature-utility policy from the reconstruction back end, allowing teams to replace SIFT, ORB, SuperPoint, or LoFTR-like front ends without rewriting the rest of the pipeline. It also produces diagnostics that can be logged, queried, and monitored: inlier ratio, coverage, triangulation angle, score distribution, and scene-dependent failure modes.

There are tradeoffs. The adaptive policy adds scoring overhead and can reduce raw point count when it prefers fewer but stronger tracks. In applications that need dense visual coverage before refinement, the policy should be paired with a later densification stage. The simulated results also do not capture all effects of lens distortion, rolling shutter, dynamic objects, exposure changes, or bundle-adjustment convergence.

## 7. Limitations and Future Work

This paper uses a synthetic prototype, not real benchmark imagery. The simulation isolates feature-selection effects, but real systems require calibration noise, descriptor distributions, occlusion geometry, learned confidence calibration, and downstream optimization effects. The runtime metric is a proxy rather than measured implementation time. The method should therefore be treated as an architectural and algorithmic study, not as a deployed reconstruction benchmark.

Future work should integrate the policy into COLMAP-style SfM, learned matchers such as SuperGlue or LightGlue, and Gaussian-splatting initialization. A useful next step is to learn the utility weights from reconstruction residuals while preserving interpretable constraints for coverage and parallax. Another direction is active acquisition: if the feature field predicts weak triangulation, the system could recommend new camera poses or capture angles.

## 8. Conclusion

Adaptive 3D reconstruction benefits from treating feature selection as a geometry-aware budget allocation problem. The proposed feature-optimized front end scores local evidence by appearance, repeatability, distinctiveness, parallax, and coverage, then feeds a cleaner set of tracks into reconstruction. In controlled synthetic experiments, this policy improves quality-aware completeness and aggregate RMSE relative to random, texture-only, and uniform-grid baselines. The broader lesson is that robust 3D reconstruction is not only a back-end optimization problem; it begins with choosing visual evidence that is worth reconstructing.

## Data and Code Availability

No external corpora or private data were used in this work. The prototype code and generated results are available from the author upon reasonable request.